\documentclass{article}

\usepackage{microtype}
\usepackage{graphicx}
\usepackage{subfigure}
\usepackage{booktabs} 

\usepackage{hyperref}

\usepackage[accepted]{icml2024}

\usepackage{amsmath}
\usepackage{amssymb}
\usepackage{mathtools}
\usepackage{amsthm}
\usepackage{siunitx}

\usepackage[capitalize,noabbrev]{cleveref}

\theoremstyle{plain}

\theoremstyle{definition}

\theoremstyle{remark}

\usepackage[textsize=tiny]{todonotes}

\begin{document}

\twocolumn[
\icmltitle{Scaling few-shot spoken word classification with generative meta-continual learning}



\icmlsetsymbol{equal}{*}

\begin{icmlauthorlist}
\icmlauthor{Louise Beyers}{equal,bytefuse}
\icmlauthor{Batsirayi Mupamhi Ziki}{equal,bytefuse}
\icmlauthor{Ruan van der Merwe}{bytefuse}
\end{icmlauthorlist}

\icmlaffiliation{bytefuse}{Bytefuse}

\icmlcorrespondingauthor{Louise Beyers}{louise@beyers.co.za}
\icmlkeywords{spoken word classification, keyword spotting, meta-learning, continual learning, meta-continual learning}

\vskip 0.3in
]



\printAffiliationsAndNotice{}  

\begin{abstract}
Few-shot spoken word classification has largely been developed for applications where a small number of classes is considered, and so the potential of larger-scale few-shot spoken word classification remains untapped. 
This paper investigates the potential of a spoken word classifier to sequentially learn to distinguish between $1000$ classes when it is given only five shots per class. 
We demonstrate that this scaling capability exists by training a model using the Generative Meta-Continual Learning (GeMCL) algorithm and comparing it to repeatedly trained or finetuned baselines. 
We find that GeMCL produces exceptionally stable performance, and although it does not always outperform a repeatedly fully-finetuned HuBERT model  nor a frozen HuBERT model with a repeatedly trained classifier head, it produces comparable performance to the latter while adapting $2000$ times faster, having been trained less than half of the data for two orders of magnitude less time.
\end{abstract}

\begin{figure*}[t!]
  \centering
  \includegraphics[width=0.85\linewidth]{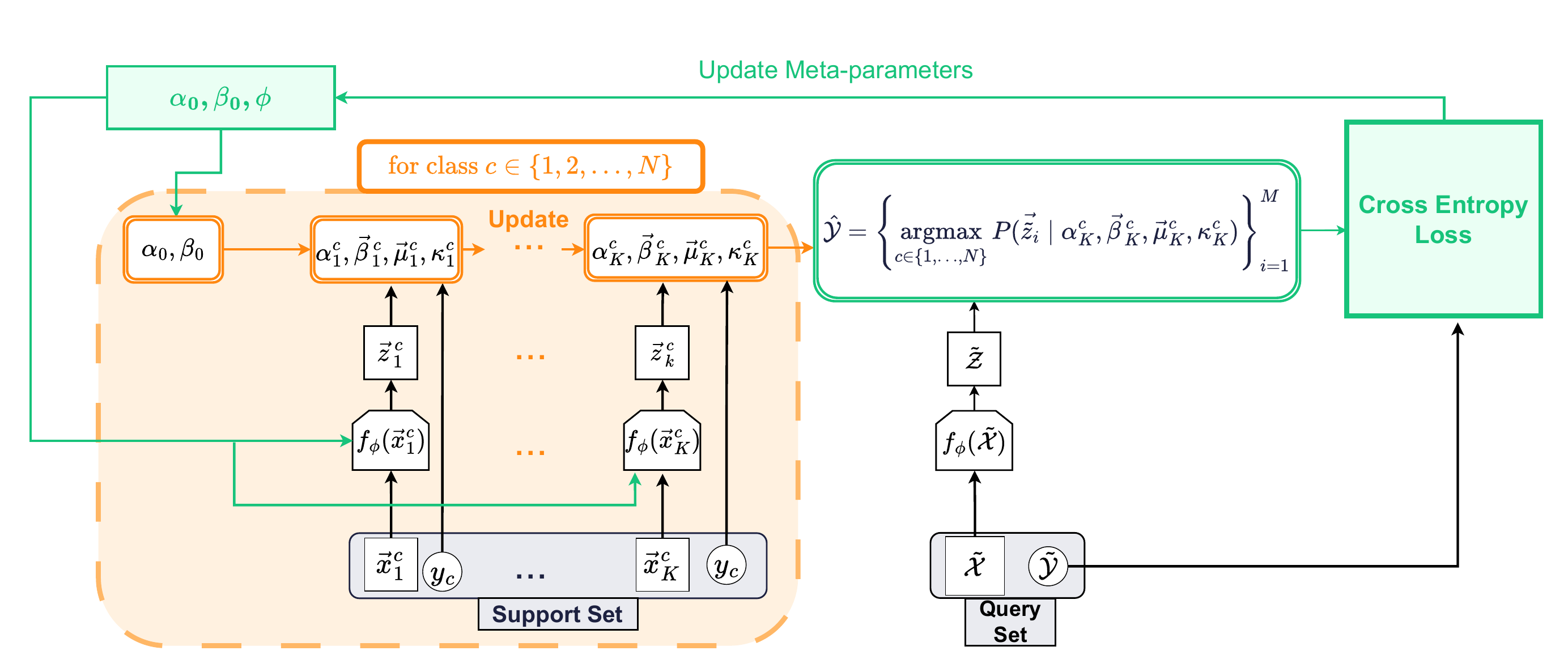}
  \caption{A single step in the meta-training procedure of GeMCL. Each task is an $N$-way-$K$-shot episode. The encoder, $f_\phi$, embeds the input. The support set is used to calculate the class statistics from the embeddings. The matrix $\tilde{\mathcal{X}}$ contains the samples from the combined query set of all classes, totalling $M$ samples, whereas the vector $\tilde{\mathcal{Y}}$ contains the corresponding labels. The model produces a vector $\hat{\mathcal{Y}}$ containing $M$ predictions. The cross-entropy loss is calculated, and used to update the meta-parameters.}
  \label{fig:gemcl_procedure}
\end{figure*}
\section{Introduction}


Spoken word classification is often researched in the context of keyword spotting (KWS) - specifically in the scenario where users are allowed to define their own keywords, and the model must adapt to the new words with only a few labelled shots \cite{2,7,5,6}. While this problem is usually formulated as an open-set classification problem \cite{8,9,10,12}, spoken word classification is generally closed-set \cite{4,5,6}; if we guarantee that all unseen words will be provided with labels from the user, then the open-set classification problem can be turned into a closed-set problem, with the addition of a continual learning aspect \cite{5}. Spoken word classification evaluation in this setting is partially driven by the requirements of edge devices \cite{13,12,7,2} and is often tested on a small number of classes at a time, since a keyword system often does not require there to be many possible classes at a time \cite{4, 10} and even at small numbers of classes, we observe limitations in performance \cite{2}.

However, if few-shot, continually learning spoken word classifiers are able to scale to large amounts of classes, the scope of the problems to which they may be applied expands significantly, from adding twenty new keywords to a keyword spotting system to potentially creating a data labelling flywheel for moderately-resourced languages.

The requirements for such a system are twofold: good performance given access to only a moderate amount of data from the same distribution, and an unprecedented ability to generalise to words not seen in the training data.

Meta-continual learning is a framework that fulfils these requirements as it lies at the intersection of meta-learning, traditionally used for few-shot classification \cite{prototypical, timothy_m_hospedales_meta-learning_2020, finn2017model}, and continual learning, and can be defined as learning how to continually learn \cite{jaehyeon_son_when_2023}. Generative Meta-Continual Learning (GeMCL) \cite{mohammadamin_banayeeanzade_generative_2021} uses a generative classifier, and is a generalisation of prototypical networks \cite{prototypical}. There is a precedent for using prototypical networks for both versions of the user-defined KWS problem \cite{12,10,9,7,4} particularly in practice because they take away the need to finetune. They have also been used to produce a model robust to numbers of shots and classes \cite{4}.

There is also a precedent for using meta-learning and meta-continual learning for few-shot continual spoken word classification to improve generalisation \cite{5,6}. Although meta-learning algorithms are often trained from scratch, they may be less data-hungry than infamously data-hungry foundation models such as HuBERT \cite{hubert} and Wav2Vec2 \cite{wav2vec2}, which would make them a more reasonable alternative when we cannot expect to rely on transfer learning from out-of-distribution data. Another alternative investigated by \cite{10} combines large self-supervised models with meta-trained projectors and classifier heads.

GeMCL isolates the class-specific parameters and is therefore immune to catastrophic forgetting \cite{mohammadamin_banayeeanzade_generative_2021}. We therefore hypothesise that it may be able to generalise and scale beyond the approaches tried so far. We thus train the meta-continual learning algorithm GeMCL \cite{mohammadamin_banayeeanzade_generative_2021} on the Multilingual Spoken Words Corpus (MSWC) \cite{mswc} dataset.

In line with the findings of \cite{simplebaseline} on appropriate baselines for continual learning, we produce baselines by selecting two adaptation strategies and applying them to the popular pre-trained (\cite{8,10,3}) HuBERT \texttt{base} model. For the first baseline we perform full finetuning of the pre-trained model, and for the second we train a classifier head on the pre-trained model. HuBERT \texttt{base} has significantly more data and compute available for training than our implementation of GeMCL, though its data source does not match the test distribution. 
GeMCL has access to data which is lower in quantity but more relevant to the task at hand. Comparing the models is therefore not only a comparison of algorithms, but of strategies: given enough data to train GeMCL but not enough data or compute to train HuBERT from scratch, is it more practical to finetune a large self-supervised model like HuBERT or is training GeMCL from scratch a viable option in terms of accuracy and compute?

Our contribution is to answer this question by comparing the two algorithms in a setting which matches practical requirements: the models must perform continual learning, they only have access to few ($5$) shots per word to do so, and they must be able to scale the number of words they can classify. 
We are the first to our knowledge to report on the application of GeMCL on speech data, as well as the first to report results for few-shot spoken word classification for as many as $1000$ classes at one time. Our contribution is also distinctive in that we test at multiple points during the continual learning process, so that we are able to report on per-class continual learning stability.

\section{Methodology}

We discuss GeMCL, our baselines and specify how we evaluate for a continual learning scenario. 

\subsection{GeMCL} 

We provide a short overview of the GeMCL algorithm; refer to \cite{mohammadamin_banayeeanzade_generative_2021} and  \cite{lee_learning_2024} for more details.

GeMCL consists of an encoder followed by a generative classifier that models the distribution of each class in the embedding space.
In a continual learning setting, new data arrives sequentially and the distribution of a class evolves as new samples are observed.
GeMCL adopts a Bayesian approach in which each class is represented by a prior distribution that gets updated to a posterior after observing new data.
Each class distribution is modelled as a Gaussian. The Gaussians collectively form a Gaussian mixture model.

For class $c$, the mean ($\vec{\mu}^c \in \mathbb{R}^d$) and precision ($\vec{\lambda}^c \in \mathbb{R}^d_+$) are modelled by Normal-Gamma distribution
\begin{equation}
P(\vec{\mu}^c, \vec{\lambda}^c \mid \theta_0) \propto \prod_{i=1}^d \mathcal{N}\big(\mu^c_i \mid \mu_0, (\kappa_0\lambda^c_i)^{-1}\big)\mathrm{Ga}\big(\lambda^c_i \mid \alpha_0, \beta_0\big),
\end{equation}
where $\theta_0 = \{\alpha_0, \beta_0, \kappa_0, \mu_0\}$ are parameters of the Normal-Gamma distribution, encoding our prior belief over the class distributions.
The initial values of the parameters represent the prior.
In practice the prior distribution for $\vec{\mu}^c$ is flat and uninformative; therefore, $\kappa_0=0$ and $\mu_0=0$.
Since the Normal-gamma is the conjugate prior to the Gaussian likelihood, the posterior remains in the same family, allowing parameters to be updated in closed form via Bayes' rule.
Assume that we receive the $n$th sample from class $c$, then the element-wise update rule for the parameters is as follows \cite{mohammadamin_banayeeanzade_generative_2021},
\begin{align}
    \kappa_n^c &= n, \label{eq:kappa} \\
    \mu_{n,i}^c &= \frac{1}{n} \sum_{i=1}^n z_i=\bar{z}_{n,i}, \label{eq:mu} \\
    \alpha_n^c &= \alpha_0 + \frac{n}{2}, \label{eq:alpha} \\
    \beta_{n,i}^c &= \beta_0 + \frac{n}{2} (\bar{\gamma}_{n,i} - \bar{z}_{n,i}^{2}), \label{eq:beta}
\end{align}
where $z_i$ is an element of the embedded input $\vec{z}\in \mathbb{R}^d$ and $\bar{\gamma}_{n,i}=\frac{1}{n}\sum_{i=1}^n z_i^2$.

These updates are class-specific, allowing previously learned class parameters to remain unaffected by new observations from different classes.
For an embedding input $\vec z$, the predicted label ${\hat{y}}$ is
\begin{equation}
    {\hat{y}} = \underset{{y}}{\rm{argmax}}\, P\left(\vec{z} | \alpha_n^{{y}}, \vec{\beta}_n^{{y}}, \vec{\mu}_n^{{y}}, \kappa_n^{{y}} \right),
\end{equation}
where $P$ is a Student's t-distribution.
GeMCL makes use of meta-learning to train the encoder to maximise its representational capacity and to obtain the prior beliefs $\beta_0$ and $\alpha_0$. 

Meta-learning can be described by two phases: meta-training and meta-testing.
Meta-training involves training the model on a distribution of tasks whereas meta-testing is evaluating whether the model can generalise to new, unseen tasks.
The parameters of the encoder and the parameters $\beta_0$ and $\alpha_0$ are optimised during meta-training and are known as the meta-parameters.

A single step in meta-training involves a model learning a task sampled from the task distribution.
Each task in the training distribution is a classification episode containing $N$ classes, with $K$ training samples for each class, which we call the support set. This is known as an $N$-way-$K$-shot episode.

During an $N$-way-$K$-shot episode the model learns one class at a time by embedding all the shots of that class and then calculating the Normal-Gamma distribution of that class by making use of equations \eqref{eq:kappa} - \eqref{eq:beta}.
Each class also has a query set, where we use cross-entropy to calculate the query set loss. The combined query set loss over each episode is then used to update the meta-parameters.
Figure \ref{fig:gemcl_procedure} illustrates the meta-training procedure of GeMCL.

During meta-testing the meta-parameters are frozen and we evaluate the model on $N$-way-$K$-shot episodes in which none of the classes were seen during meta-training.

\subsection{HuBERT baselines}

The HuBERT \texttt{base} model is chosen as the starting point for our baselines as it is known to produce strong representations of speech \cite{8,10,3}. We choose two methods of adapting the HuBERT \texttt{base} encoder - full finetuning and training a classifier head and projector on a frozen backbone - to contextualise the performance of GeMCL.

To fully finetune HuBERT, we allow updates to all parameters during finetuning. This affords maximum flexibility and prevents the model from being limited by representations that are optimised for a different distribution, and so provides a measure of how well we can transfer knowledge in a scenario with full flexibility. However, full finetuning is known to be sensitive to the choice of hyperparameters, and so it trades stability for performance.

Our other baseline trains a classifier head on top of a frozen HuBERT model, to test whether the combination of a strong encoder and weak classifier can perform continual learning. This baseline tests how well a customisable classifier head can perform given a strong encoder. Although this affords less flexibility, it is far cheaper and more robust, since no previously-trained parameters are updated.

\subsection{Evaluating for continual learning}

Many studies concerning class-incremental continual learning \cite{incrementallearning} test the performance of a model only once during the continual learning process: at the end. Catastrophic forgetting can then be detected by comparing the order in which a class is learned to the performance of the model on the class.

However, in a practical continual learning setting, a model might be updated and deployed multiple times. This motivates reporting how the accuracy of a model changes over time. We therefore test all models at intermediate points of the continual learning process on the classes they have seen by those points.

\begin{figure}[t]
  \centering
  \includegraphics[width=\linewidth]{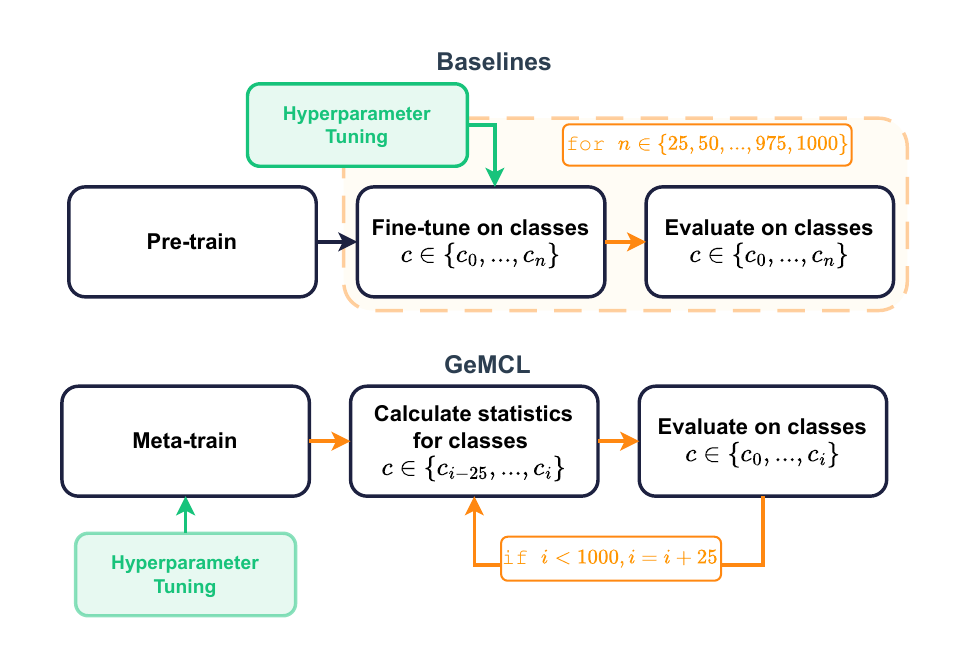}
  \caption{A comparison of the training and evaluation flows of the baselines and GeMCL}
  \label{fig:flow compare}
\end{figure}

Figure \ref{fig:flow compare} illustrates how evaluation of the models differs. Since GeMCL can ingest support sets for different classes independently, there is no additional learning cost associated with evaluating at different stages of learning.
However, our baselines must be finetuned from scratch each time new classes are considered. There can be an additional hyperparameter tuning cost associated with this. The baselines therefore do not provide a comparison in terms of adaptability, but they instead provide a measure of the possible performance that can be achieved if you allow training on all classes at once, instead of sequentially.

\section{Experimental setup}
We describe the data, architectures, training methods and evaluation methods used to compare GeMCL to a fully finetuned HuBERT (full FT) and frozen HuBERT model with a finetuned classifier head and projector (CH).

\subsection{Data}
We use English data from the Multilingual Spoken Words corpus (MSWC) \cite{mswc} dataset, which provides one-second-long labelled audio segments of individual words. A train/dev/test split is provided which splits on samples, and has all words in all splits. We only use the test and train splits. After filtering for valid links and words with at least five test and train recordings in the original dataset's split, we are left with $12736$ words. We split these words in a 70-30 ratio for meta-training and meta-testing, respectively. We use five shots per word for training, validation and testing, sampled from the training, training and test splits respectively.

\subsection{GeMCL model and training}
Our GeMCL implementation, inspired by Lee et al.'s implementation \cite{lee_learning_2024}, employs a 12-layer-12-head transformer encoder backbone which takes mel-frequency cepstral coefficients (MFCCs) as input, producing a model with \num{85066756} parameters. MFCCs are extracted from waveforms sampled at 16{kHz}, using a frame length of 25{ms}, a frame shift of 10{ms}, and 40 mel filterbanks. We retain the first 13 cepstral coefficients. 

We train GeMCL on the $8915$ reserved meta-train words of the dataset for $5000$ steps on batches of $16$ $25$-way-$5$-shot episodes. GeMCL is therefore fed $10$ million samples - about $2778$ hours - though not all samples seen may be unique. We simulate our sampling procedure for $500$ iterations, giving an average of about $1715842$ unique samples: about $477$ hours of training data.

\begin{figure*}[t!]
  \centering
  \includegraphics[width=\linewidth]{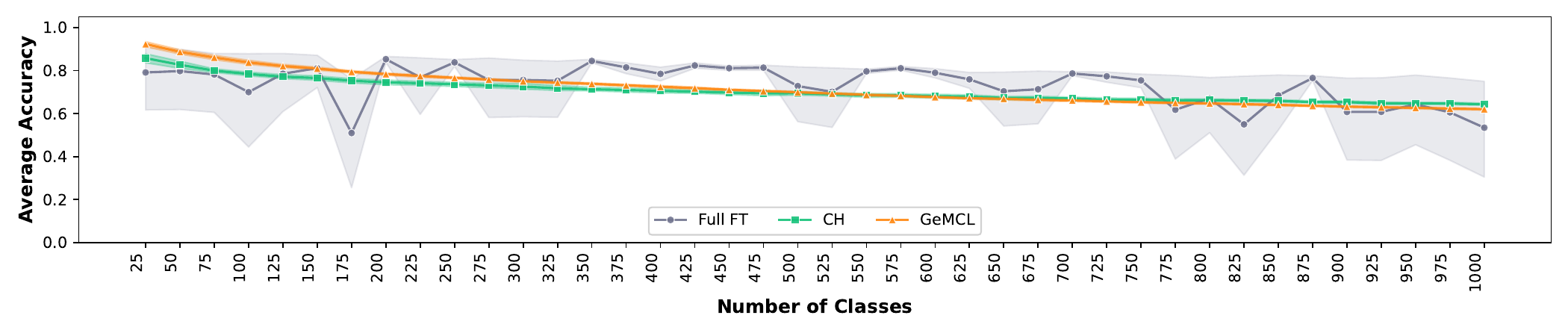}
  \caption{The average accuracy of GeMCL, full FT and the CH models with $95\%$ confidence intervals.}
  \label{fig:money_shot_hopefully}
\end{figure*}

\subsection{Baseline models and training}
For both baselines, we use the pre-trained HuBERT \texttt{base} model with an added projector and classifier head. The model parameters range from \num{94575001} (for $25$ classes) to \num{94825576} (for $1000$ classes). HuBERT \texttt{base} was pre-trained on $960$ hours of data from LibriSpeech \cite{librispeech}. The baselines take raw waveform as input, padded with zeros to the maximum length of a sample in a batch. We train both baselines with a cross-entropy objective for $200$ epochs if the number of classes is below $300$, and $500$ epochs if the number of classes is above $300$. The batch size is $32$ and we use the AdamW \cite{adamw} optimiser for both models. We perform training runs, each starting from the HuBERT \texttt{base} checkpoint, for sets of classes in the range $[25, 1000]$ where the number of classes is incremented by $25$, as illustrated by Figure \ref{fig:flow compare}. For each training run, the dimension of the classifier head output is the number of classes considered at that point. The learning rate for the CH model is 3e-4 and for the full FT model the learning rate is dependent on the number of classes being learned. The only other difference between the baselines is that for the CH model, the only trainable parameters are the projector and classifier head, whereas for the full FT model all parameters are trainable.

\subsection{Evaluation}

We evaluate all models by exposing them to a support set of five shots per class (for GeMCL this means calculating class statistics, for the baselines this means finetuning, see Figure \ref{fig:flow compare}) for a number of classes and then recording their accuracy on five unseen shots (the query set) from each of those classes.
We do so for 25 classes, then add 25 classes and repeat the process until we reach 1000 classes. We repeat this for ten different selections of 1000 classes from the meta-test set of words, each of which we call an episode. We report on the average accuracy over words and episodes, as well as the average change in accuracy per-word between consecutive points in the continual learning process.

We also report the hours spent on tuning, training and few-shot adaptation for each model. 

\section{Results}

Figure \ref{fig:money_shot_hopefully} illustrates the accuracy of GeMCL compared to the full FT and CH models. The full FT baseline is able at most stages of continual learning to outperform CH and GeMCL, but the fluctuation in the confidence intervals for full FT show that it is not very stable. This observation is further supported by Table \ref{tab:accuracy_volatility}, which shows that the percentage accuracy of full FT on a word between two stages of the continual learning process changes by about 25{\%} on average.

\begin{table}[h]
  \caption{Per-word volatility of classification accuracy. Accuracy in \%, volatility as the average absolute amount that accuracy changes between consecutive continual learning steps.}
  \label{tab:accuracy_volatility}
  \centering
  \begin{tabular}{ r r r }
    \toprule
    \multicolumn{1}{c}{\textbf{GeMCL}} & \multicolumn{1}{c}{\textbf{CH}} & \multicolumn{1}{c}{\textbf{Full FT}} \\
    \midrule
    $0.48$ & $7.13$ & $24.55$   ~~~             \\
    $\pm 3.32$ & $\pm 13.63$ & $\pm 39.25$   ~~~               \\
    \bottomrule
  \end{tabular}

\end{table}

On the other hand, GeMCL produces the most stable per-word accuracy, fluctuating less on average by at least one order of magnitude than CH and full FT. GeMCL outperforms CH for all increments of classes below $450$, and underperforms CH at $700$ classes, as well as all increments of classes above $750$ (significantly, according to a two-sided Mann-Whitney U test with $\alpha=0.05$).

Despite only ever being trained to distinguish $25$ classes at a time, GeMCL only underperforms $1000$-way-finetuned CH by about $2\%$.

Further, GeMCL takes less time for training, tuning and few-shot adaptation. We report the time taken for each of these in Table \ref{tab:compute}.

\begin{table}[th]
  \caption{Time taken in hours on a single GPU - the HuBERT numbers are calculated from $100$k steps taking approximately $9.5$ hours on 32 GPUs. The hours reported are cumulative - the adaptation time is taken over all shots ingested.}
  \label{tab:compute}
  \centering
  \begin{tabular}{ l  r r r }
    \toprule
    & \multicolumn{1}{c}{\textbf{GeMCL}} & \multicolumn{1}{c}{\textbf{CH}} & \multicolumn{1}{c}{\textbf{Full FT}} \\
    \midrule
    \footnotesize{\textbf{Meta- / Pre-train}} & $27.55$ & $\sim 1976$ & $\sim 1976$   ~~~               \\
    \footnotesize{\textbf{Hyperparameter search}}  & $15.29$ & $40.26$ & $70.00$   ~~~         \\
    \footnotesize{\textbf{Few-shot adaptation}} & $0.06$ & $124.25$ & $185.96$   ~~~               \\
    \bottomrule
  \end{tabular}

\end{table}

Table \ref{tab:compute} shows how much time was used, given a single GPU as a resource, to train and hyperparameter tune the three models, and to adapt them to the support set. The time used for the baselines for finetuning is exacerbated by two things: first, finetuning started from scratch each time the number of classes was incremented, and second, validation was performed more frequently than necessary. Even so, in comparison to the time taken for GeMCL to calculate prototypes, the baselines took very long to adapt.

\section{Conclusion}

We compared the generative meta-continual learning algorithm, GeMCL, trained from scratch on about $477$ hours of labelled spoken word data, to two baselines based on the HuBERT \texttt{base} model, originally trained in an unsupervised way on $960$ hours of speech data. 
We evaluated all models on 5-shot spoken word classification tasks, scaling up to $1000$ classes in a continual learning setting.
GeMCL was able to achieve accuracy within $3\%$ of the most practically viable baseline model on $1000$ classes, while being a true continual learner - in other words, it requires no retraining when new words arrive, rather a closed-form update to word class statistics.
Additionally, GeMCL's accuracy on any given word remains remarkably stable as new words are added, in contrast to the HuBERT baselines, which produce unstable per-word performance; GeMCL is therefore more suitable and predictable for real world deployment.  
These findings support the viability of GeMCL to be used for spoken word classification in a continual learning setup where  the number of classes to be learned scales. Future work could investigate the data efficiency of GeMCL on lower resource languages to expand these findings past just English systems. 

\bibliography{bib}
\bibliographystyle{icml2024}

\end{document}